\documentclass[conference]{IEEEtran}
\IEEEoverridecommandlockouts
\usepackage[noadjust]{cite}
\usepackage{amsmath,amssymb,amsfonts}
\usepackage{algorithmic}
\usepackage{graphicx}
\usepackage{textcomp}
\usepackage{xcolor}
\usepackage{amsmath}
\usepackage{algorithm}
\usepackage{algorithmic}
\usepackage{tabularx}
\usepackage{multirow}
\usepackage{booktabs}
\usepackage{float}
\usepackage{url}
\usepackage{hyperref}
\def\BibTeX{{\rm B\kern-.05em{\sc i\kern-.025em b}\kern-.08em
    T\kern-.1667em\lower.7ex\hbox{E}\kern-.125emX}}
\begin{document}

\title{MuonAll: Muon Variant for Efficient Finetuning of Large Language Models\\
\thanks{Computational Linguistics Lab, Department of Computer Engineering, Pune Institute of Computer Technology}
}

\author{\IEEEauthorblockN{Saurabh Page}
\IEEEauthorblockA{
% \textit{dept. name of organization (of Aff.)} \\
Pune, India \\
saurabhpage1@gmail.com}
\and
\IEEEauthorblockN{Advait Joshi}
\IEEEauthorblockA{\textit{Computer Engineering} \\
\textit{Pune Institute of Computer Technology}\\
Pune, India \\
advaitkjoshi@gmail.com}
\and
\IEEEauthorblockN{S. S. Sonawane}
\IEEEauthorblockA{\textit{Computer Engineering} \\
\textit{Pune Institute of Computer Technology}\\
Pune, India \\
sssonawane@pict.edu}
}

\maketitle

\begin{abstract}
Muon optimizer has demonstrated robust results in pretraining of language models but its performance in finetuning of existing public pretrained models is not yet explored. Currently, Muon is used along  with AdamW introducing a scope of improvement for adopting all parameters inside Muon. We introduce MuonAll, which incorporates all the parameters inside Muon by transforming into 2D matrices. We conduct extensive finetuning experiments across publicly available language models with model sizes upto half billion parameters. Muon and MuonAll perform at par with AdamW across major benchmarks, highlighting their effectiveness as alternative optimizers. We open-source the distributed implementations of Muon and MuonAll, available at \url{https://github.com/Saurabh750/optimizer}
\end{abstract}


\section{Introduction}
The natural language processing (NLP) field has been rapidly advancing  with the recent research efforts contributed by \cite{b10}, \cite{b11}, \cite{b12}. With the fast evolving language model space, the requirement of computational power has been ever increasing and there is a pressing need for finding efficient methods for model training. One such approach for compute efficient training is choosing the right optimizer. Adam (\cite{b13}) and AdamW (\cite{b14}) have become primary choices for pretraining and finetuning stages of language model training. 

To improve training efficiency, different optimization algorithms have been proposed (\cite{b15}, \cite{b1}, \cite{b16}, \cite{b17}, \cite{b18}, \cite{b19}, \cite{pooladzandi2024curvature}, \cite{li2022multi}, \cite{zhao2024deconstructing}, \cite{pethick2025stable}). Of these, Muon introduced by \cite{b1} performs othogonalization on momentum of matrix parameters followed by updation of matrix parameters. It shows promising results compared to AdamW in pretraining of small language models. Muon achieves target loss value in less number of tokens compared to AdamW. \cite{b2} shows that by performing some modifications in Muon architecture, it can be scaled for large language model training. The paper briefly compares Muon-AdamW performance on supervised finetuning (SFT) task on public pre-trained models. The paper mentions possibility of inclusion of all parameters into Muon.

In this paper, we provide a detailed analysis of Muon-AdamW performance on SFT task on small language models (SLMs). We explore the direction of incoporating all model parameters inside Muon framework by modifying the Muon architecture.

\section{Background}
% Muon Optimizer (Jordan et al., 2024) — Section 2.1 Background

\noindent
Muon (MomentUm Orthogonalized
by Newton-Schulz) is  a matrix orthogonalization based optimizer specially designed for two-dimensional parameters of hidden layers in a neural network \cite{b1}.
At iteration \(t\), given current weight \(\mathbf{W}_{t-1}\), momentum \(\mu\), learning rate \(\eta_t\), and objective \(\mathcal{L}_t\), its update rule can be stated as:
\[
\mathbf{M}_t \;=\; \mu\,\mathbf{M}_{t-1} \;+\; \nabla \mathcal{L}_t\!\left(\mathbf{W}_{t-1}\right) \tag{1}
\]
\[
\mathbf{O}_t = \mathrm{Newton{-}Schulz}(\mu\,\mathbf{M}_t + \nabla \mathcal{L}_t\!\mathbf{W}_{t-1}) \, \tag{2}
\]
\[
\mathbf{W}_t \;=\; \mathbf{W}_{t-1} \;-\; \eta_t\,\mathbf{O}_t \tag{3}
\]

\noindent
Here, \(\mathbf{M}_t\) is the momentum of the gradient at iteration \(t\), with \(\mathbf{M}_0 = \mathbf{0}\). In Equation \( (2) \), Newton--Schulz iteration (\cite{bernstein2024oldoptimizer}) is adopted to approximately compute \(\left(\mathbf{M}_t \mathbf{M}_t^{\top}\right)^{-1/2}\mathbf{M}_t\). In equation \( (2) \), Nesterov-style momentum is passed to the NS iterations. \cite{b1} reports that Nesterov momentum gives better empirical results than standard SGD-momentum. 

\noindent \(\left(\mathbf{M}_t \mathbf{M}_t^{\top}\right)^{-1/2}\mathbf{M}_t\), the key term involved in Muon's operation whitens the rows of \(\mathbf{M}_t\). Whitening operation normalizes the variance along each direction and decorrelates updates. It ensures momentum update doesn't favor redundant directions, which is a common inefficiency in SGD or Adam. If we have access to Singular Value Decomposition (SVD) of \(\mathbf{M}_t\) i.e.
\[
\mathbf{M}_t \;= \; \mathbf{U}\mathbf{\Sigma}\mathbf{V}^{\top}
\]
whitening looks like:
\[
\left(\mathbf{M}_t \mathbf{M}_t^{\top}\right)^{-1/2}\mathbf{M}_t \;=\; \mathbf{U}\mathbf{V}^{\top},
\]
This transformation retains directionality in the form of \(\mathbf{U}\mathbf{V}^{\top}\) but removes scale(i.e. singular values which are stored in \(\mathbf{\Sigma}\)) leading to orthogonal updates. Calculation of SVD increases Muon's wallclock time. To tackle this, Newton-Schulz iteraton is used. One iteration of Newton-Schulz looks like this: 
\begin{align*}
G’ &:= aG + b(GG^\top)G + c(GG^\top)^2G \\
&= (aI + b(GG^\top) + c(GG^\top)^2)G \\
&= (aI + bUS^2U^\top + cUS^4U^\top)USV^\top \\
&= U(aS + bS^3 + cS^5)V^\top \\
&= U \varphi(S) V^{\top}
\end{align*}

Here, \( G = M_t / ||M_t||_F \) where \( ||M_t||_F \) is the Frobenius Norm. Applying \(N\) steps of NS iteration gives us the following output \( U\varphi^{N}(S)V^{\top} \). To make sure the output converges to \( UV^{\top} \), the values of coefficients \(a,b,c\) are tuned and set to (3.4445, 4.7750, 2.0315) in the original Muon implementation such that the \( \varphi^N(x) \to 1 \)

Convergence of Muon under different conditions are analyzed and discussed in (\cite{b6}, \cite{b7}, \cite{b8}, \cite{b9}). J. Li and M. Hong analyzed Muon as a specialized steepest descent method whose update direction solves a quadratic model under a spectral-norm constraint, and establishes convergence guarantees for both a Nesterov-style and a heavy-ball-style variant of the algorithm \cite{b6}. Rigorous convergence proofs for Muon across four realistic variants (with/without Nesterov momentum and with/without decoupled weight decay), show the standard configuration with both momentum and weight decay attains the tightest theoretical bounds and fastest empirical convergence \cite{b7}.
Convergence guarantees for Muon in nonconvex settings are developed and its rates are compared with Gradient Descent, identifying structural Hessian conditions under which Muon is provably faster \cite{b8}. 
L. Chen, J. Li, and Q. Liu gave a theoretical framing of Muon by placing it inside the Lion‑\(\mathcal{K}\) optimizer family with \(\mathcal{K}\) equal to the nuclear norm, implying that Muon with decoupled weight decay implicitly solves a spectral‑norm constrained optimization problem on weight matrices in \cite{b9}.

An adaptive variant of Muon is proposed in \cite{b4} that combines element‑wise second‑moment scaling with orthogonalized updates to improve stability and efficiency for large‑scale training. Empirical results report that AdaMuon maintains Muon’s stability and can surpass Adam by over 40\% training efficiency at scale across models and learning‑rate regimes, indicating practical gains in convergence speed without added complexity. A study of DiLoCo’s inner optimizer and communication compressibility \cite{b5}, shows that replacing AdamW with Muon plus error‑feedback enables aggressive compression of the communicated deltas to 2‑bit with negligible loss while preserving memory complexity. 

Essential AI \cite{b3} evaluated Muon’s practical pretraining efficiency with a focus on compute–time and data–batch tradeoffs, showing that it expands the Pareto frontier over AdamW by maintaining data efficiency at very large batch sizes while staying computationally lightweight for modern hardware. The study also demonstrates that Muon can be paired with maximal update parameterization (muP) to transfer hyperparameters across widths via a simple telescoping calibration procedure with modest overhead, validated up to ~4B-parameter models and across data/architecture ablations.
Empirically, Muon achieves target losses faster at a fixed device count and achieves similar losses with fewer tokens at fixed wall-clock, explicitly improving the resources–time frontier relative to AdamW for pretraining transformers up to multi-billion scale. 
J. Liu has provided a comprehensive study on the scalability of Muon in LLM training \cite{b2}. The study demonstrates that Muon can effectively replace AdamW as the standard optimizer for large-scale LLM training. The authors scale up Muon to large language models by doing the following: 1) Adding standard AdamW weight decay mechanism to Muon 2) carefully adjusting per-parameter update scale. The authors opensource Moonlight model(a 3B/16B-parameter MoE model trained on 5.7 trillion tokens) and intermediate training checkpoints. The authors show that Muon performs on par with AdamW in SFT on a public pretrained model. The paper suggests several promising directions for future research that include incorporating all parameters into the Muon framework and developing approaches to better understand and address the pretraining–finetuning mismatch.

\section{MuonAll}
Currently, Muon works only for 2-dimensional parameters. AdamW optimizer is used for the remaining parameters. To remove the dependency on AdamW and with the aim of incorporating all the parameters in a single optimizer, we introduce MuonAll, modified version of Muon. MuonAll can be used as a single optimizer for all parameters including 1D parameters. MuonAll is defined as follows:
\begin{algorithm}
\caption{MuonAll}
\begin{algorithmic}[1]
\REQUIRE Learning rate $\eta$, momentum $\mu$
\STATE Initialize $B_0 \gets 0$
\FOR{$t = 1, \ldots$ n}
    \STATE Compute gradient $G_t \gets \nabla_\theta \mathcal{L}_t(\theta_{t-1})$
    \IF{paramater is one-dimensional} 
    \STATE Convert parameter into a diagonal-matrix with elements along the main diagonal 
    \ENDIF
    \STATE $B_t \gets \mu B_{t-1} + G_t$
    \STATE $O_t \gets \text{NewtonSchulz5}(G_t + \mu B_t)$
    \IF{paramater is one-dimensional} 
    \STATE Convert diagonal matrix back into one-dimensional parameter 
    \ENDIF
    \STATE Update parameters $\theta_t \gets \theta_{t-1} - \eta O_t$
\ENDFOR
\RETURN $\theta_t$
\end{algorithmic}
\end{algorithm}
MuonAll architecture has two additional steps as compared to Muon. Before calculating momentum, if the parameter is one-dimensional, we convert it to a diagonal matrix with all the elements along the main diagonal. We calculate momentum and pass all the parameters through Newton-Schulz iterations. In the second additional step, the diagonal matrix parameters from step 1 are converted back to one-dimensional parameters.

We conduct model finetuning experiments and compare the performance of MuonAll with Muon and AdamW in Section \ref{sec:Exp}.

\section{Experiments}
\label{sec:Exp}
We perform supervised finetuning(SFT) experiments on 3 base models namely Qwen2-0.5B \cite{qwen2}, SmolLM2-360M \cite{allal2025smollm2} and GPT2 medium \cite{radford2019language}. We randomly pick 400k samples from publicly available OpenOrca dataset \cite{OpenOrca}. To maintain consistency, we make sure the same dataset samples are used across experiments. All the experiments are performed for 2 epochs with a maximum model sequence length of 1024. For each optimizer we use a linear learning-rate schedule that decays from an initial rate to \(10^{-7}\) at the end of epoch 2. The experiments were performed on 2 NVIDIA RTX A6000 GPUs.

\noindent\textbf{Evaluation Benchmarks:} Our evaluation uses several benchmarks to assess language model in different ways:
\begin{itemize}
    \item \textbf{Knowledge QA}: MMLU \cite{hendrycks2021mmlu}, ARC‑Easy \cite{b21}(science focus).
    \item \textbf{Commonsense}: HellaSwag \cite{zellers2019hellaswag}, Winogrande \cite{sakaguchi2019winogrande}, PIQA \cite{b20}.
    \item \textbf{Reading/factuality}: BoolQ \cite{b22}.
    \item \textbf{Math reasoning}: GSM8K \cite{cobbe2021gsm8k}.
    \item \textbf{Science composition}: OpenBookQA \cite{mihaylov2018openbookqa}(claiming knowledge composition/RAG)
    \item \textbf{LM quality}: LAMBADA \cite{lambadaopenai} (perplexity/LM behavior rather than task QA)
\end{itemize}

\subsection{SFT on Qwen2-0.5B}
\begin{figure}[H]
    \centering
    \includegraphics[width=0.5\textwidth]{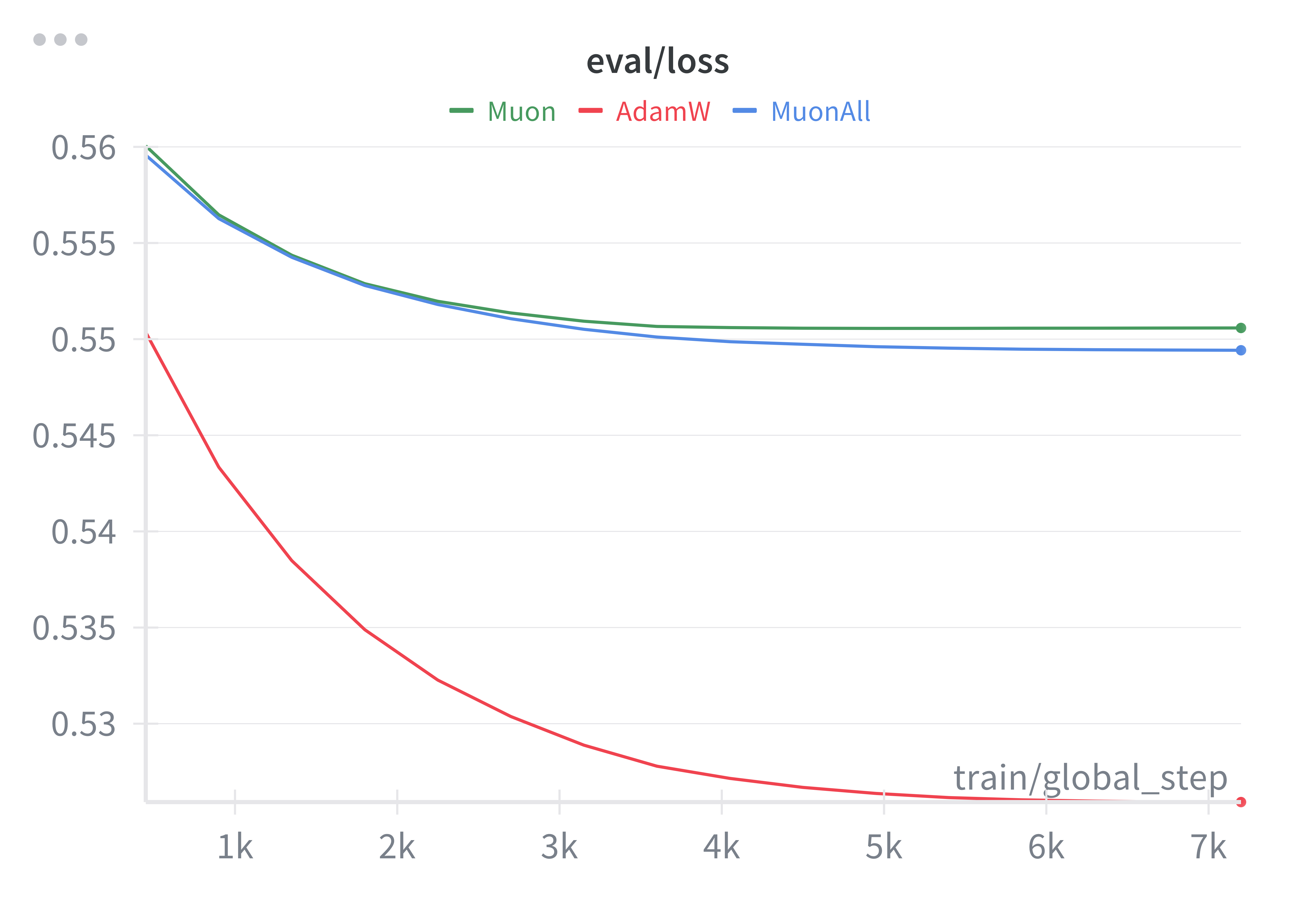} % adjust width as needed
    \caption{Validation loss on Qwen 0.5B}
    \label{fig:qwen-val}
\end{figure}

SFT experiments are performed using AdamW, MuonAll and Muon optimizer. For all three experiments, an effective batch size of 48 (batch size: 12, gradient accumulation: 4) is used. Following initial learning rates are used for AdamW and MuonAll: \(2\) x \(10^{-5}\), \(5\) x \(10^{-5}\) respectively. For Muon, two learning rates need to be set: \(5\) x \(10^{-5}\)(for matrix parameters) and \(2\) x \(10^{-5}\) (for non-matrix parameters).
The training and validation loss for each experiment can be found in table \ref{tab:val-results}. Validation loss curves can be found in figure \ref{fig:qwen-val}. The finetuned versions are evaluated over the benchmarks. The benchmarking scores can be found in table \ref{tab:qwen-bench}.

\begin{figure}[H]
    \centering
    \includegraphics[width=0.5\textwidth]{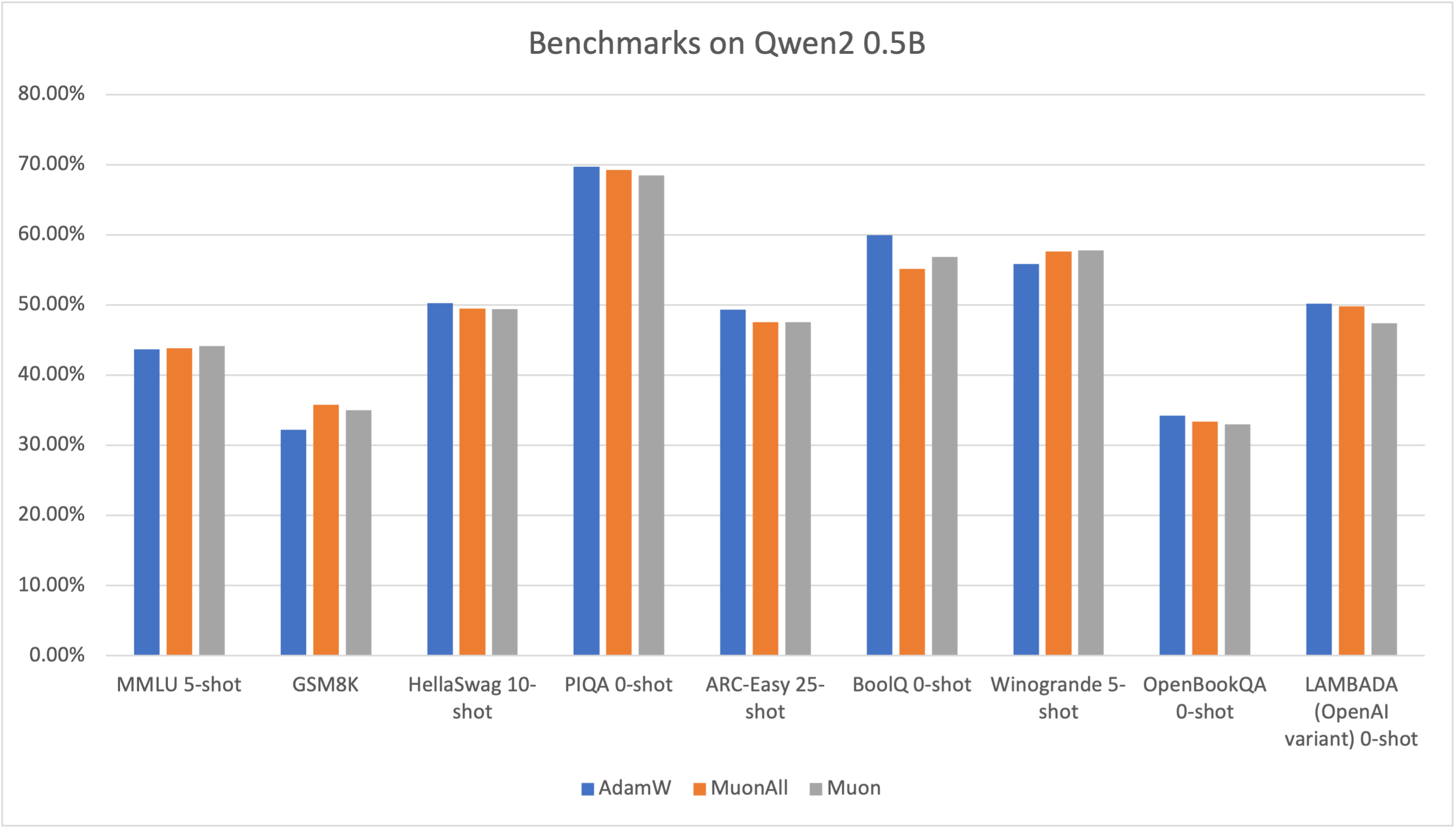} % adjust width as needed
    \caption{Benchmarks on Qwen2 0.5B}
    \label{fig:qwen-hist}
\end{figure}

The performance of Muon and MuonAll is comparable with AdamW. In the benchmarks where AdamW outperforms Muon; MuonAll outperforms Muon as well. Muon versions perform better than AdamW in major benchmarks: MMLU and GSM8K. The figure \ref{fig:qwen-hist} gives a good visualization of the benchmarking scores.

\begin{table}[H]
\centering
\renewcommand{\arraystretch}{1.2}
\caption{Benchmark evaluation results of Qwen2-0.5B}
\setlength{\tabcolsep}{4pt}
\resizebox{\linewidth}{!}{%
\begin{tabular}{l c c c}
\hline
\textbf{Benchmark} & \textbf{AdamW} & \textbf{MuonAll} & \textbf{Muon} \\
\hline
MMLU 5-shot              & 43.64\%  & 43.82\%  & \textbf{44.17\%}  \\
GSM8K                    & 32.22\%  & \textbf{35.78\%}  & 35.03\%  \\
HellaSwag 10-shot        & \textbf{50.22\%} & 49.51\% & 49.39\% \\
PIQA 0-shot              & \textbf{69.70\%} & 69.26\% & 68.50\% \\
ARC-Easy 25-shot         & \textbf{49.33\%} & 47.56\% & 47.56\% \\
BoolQ 0-shot             & \textbf{59.97\%} & 55.11\% & 56.85\% \\
Winogrande 5-shot        & 55.80\% & 57.62\% & \textbf{57.77\%} \\
OpenBookQA 0-shot        & \textbf{34.20\%} & 33.40\% & 33.00\% \\
LAMBADA (OpenAI variant) 0-shot & \textbf{50.15\%} & 49.76\% & 47.41\% \\
\hline
\end{tabular}%
}
\label{tab:qwen-bench}
\end{table}

\subsection{SFT on SmolLM2-360M}
\begin{figure}[H]
    \centering
    \includegraphics[width=0.5\textwidth]{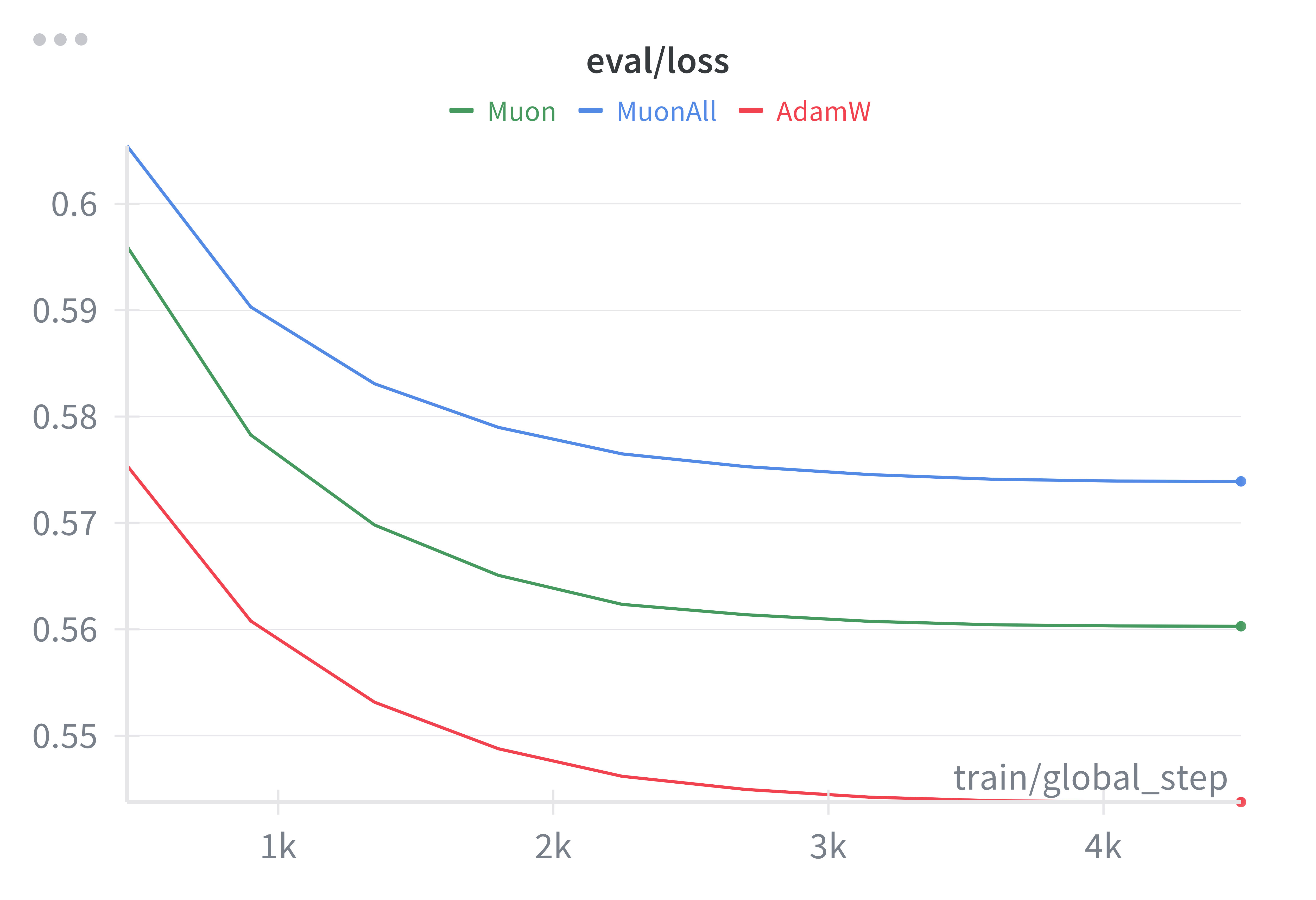} % adjust width as needed
    \caption{Validation loss on SmolLM 360M}
    \label{fig:smolLM-val}
\end{figure}

Experiments are performed using the following optimizers: AdamW, MuonAll and Muon. An effective batch size of 80 (batch size: 20, gradient accumulation: 4) is used. Following initial learning rates are used for AdamW and MuonAll: \(2\) x \(10^{-4}\), \(8\) x \(10^{-4}\) respectively. For Muon, two learning rates need to be set: \(7\) x \(10^{-4}\)(for matrix parameters) and \(2\) x \(10^{-4}\) (for non-matrix parameters).
The training and validation loss for each experiment can be found in table \ref{tab:val-results}. Validation loss curves can be found in figure \ref{fig:smolLM-val}. The finetuned versions are evaluated over the benchmarks. The benchmarking scores can be found in table \ref{tab:smolLM-bench}.

\begin{figure}[H]
    \centering
    \includegraphics[width=0.5\textwidth]{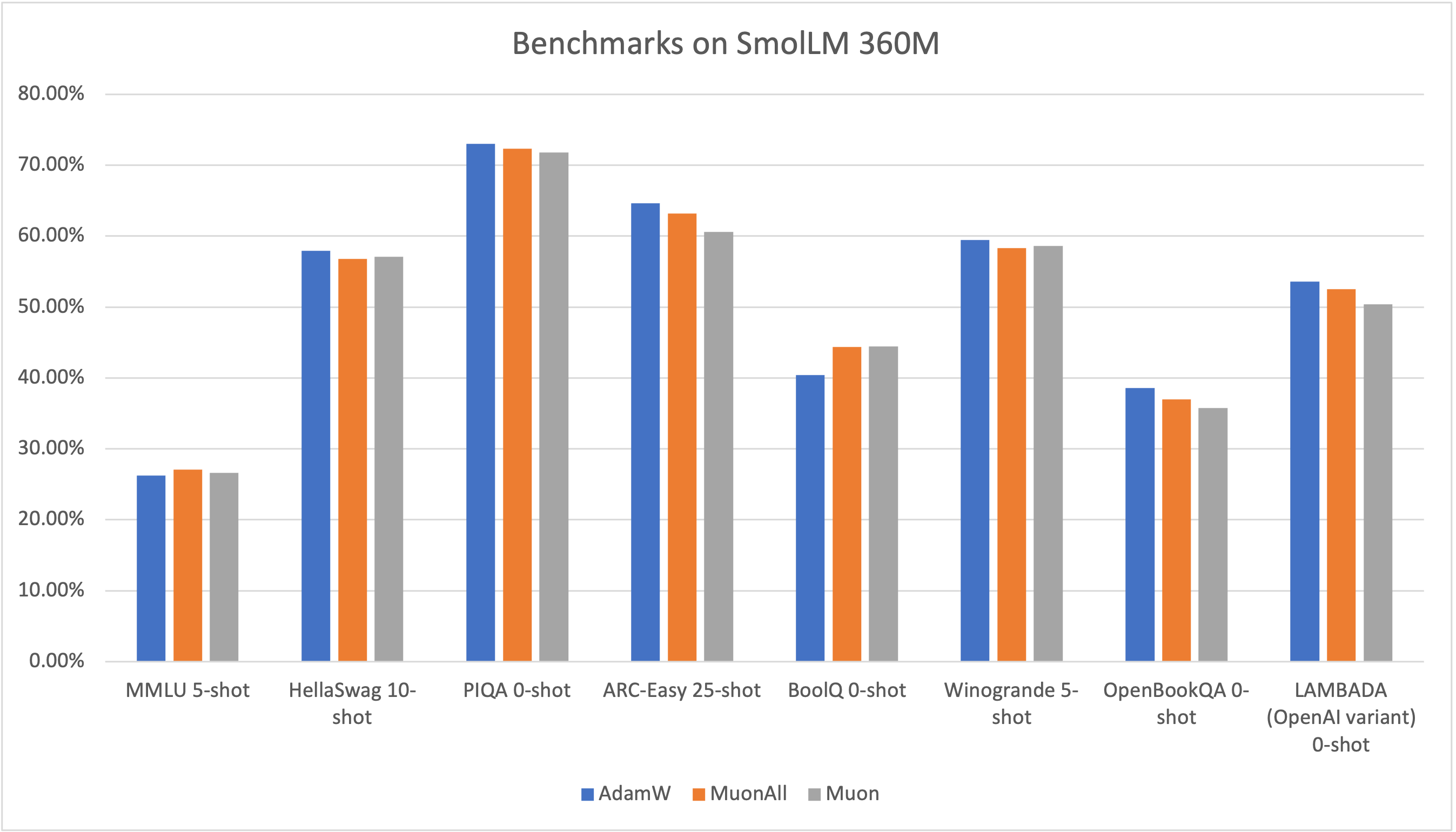} % adjust width as needed
    \caption{Benchmarks on SmolLM 360M}
    \label{fig:smolLM-hist}
\end{figure}

The performance of Muon and MuonAll is comparable with AdamW. In the benchmarks where AdamW surpasses Muon; MuonAll surpasses Muon as well. Muon versions perform better than AdamW in major benchmark MMLU. The figure \ref{fig:smolLM-hist} gives a good visualization of the benchmarking scores.

\begin{table}[H]
\centering
\renewcommand{\arraystretch}{1.2}
\caption{Benchmark evaluation results of SmolLM2-360M}
\setlength{\tabcolsep}{4pt}
\resizebox{\linewidth}{!}{%
\begin{tabular}{l c c c}
\hline
\textbf{Benchmark} & \textbf{AdamW} & \textbf{MuonAll} & \textbf{Muon} \\
\hline
MMLU 5-shot              & 26.26\% & \textbf{27.08\%} & 26.65\% \\
HellaSwag 10-shot        & \textbf{57.95\%} & 56.80\% & 57.09\% \\
PIQA 0-shot              & \textbf{73.01\%} & 72.36\% & 71.76\% \\
ARC-Easy 25-shot         & \textbf{64.65\%} & 63.17\% & 60.61\% \\
BoolQ 0-shot             & 40.40\% & 44.34\% & \textbf{44.46\%} \\
Winogrande 5-shot        & \textbf{59.43\%} & 58.33\% & 58.64\% \\
OpenBookQA 0-shot        & \textbf{38.60\%} & 37.00\% & 35.80\% \\
LAMBADA (OpenAI variant) 0-shot & \textbf{53.62\%} & 52.53\% & 50.42\% \\
\hline
\end{tabular}%
}
\label{tab:smolLM-bench}
\end{table}

\subsection{SFT on GPT2-medium}
AdamW, MuonAll and Muon are used in the SFT experiments for GPT2-medium model. An effective batch size of 60 (batch size: 15, gradient accumulation: 4) is used. Following initial learning rates are used for AdamW and MuonAll: \(5\) x \(10^{-4}\), \(9\) x \(10^{-4}\) respectively. For Muon, two learning rates need to be set: \(9\) x \(10^{-4}\)(for matrix parameters) and \(5\) x \(10^{-4}\) (for non-matrix parameters).
The training and validation loss for each experiment after two epochs can be found in table \ref{tab:val-results}.The finetuned versions are evaluated over the benchmarks. The benchmarking scores can be found in table \ref{tab:gpt2-bench}.

\begin{figure}[H]
    \centering
    \includegraphics[width=0.5\textwidth]{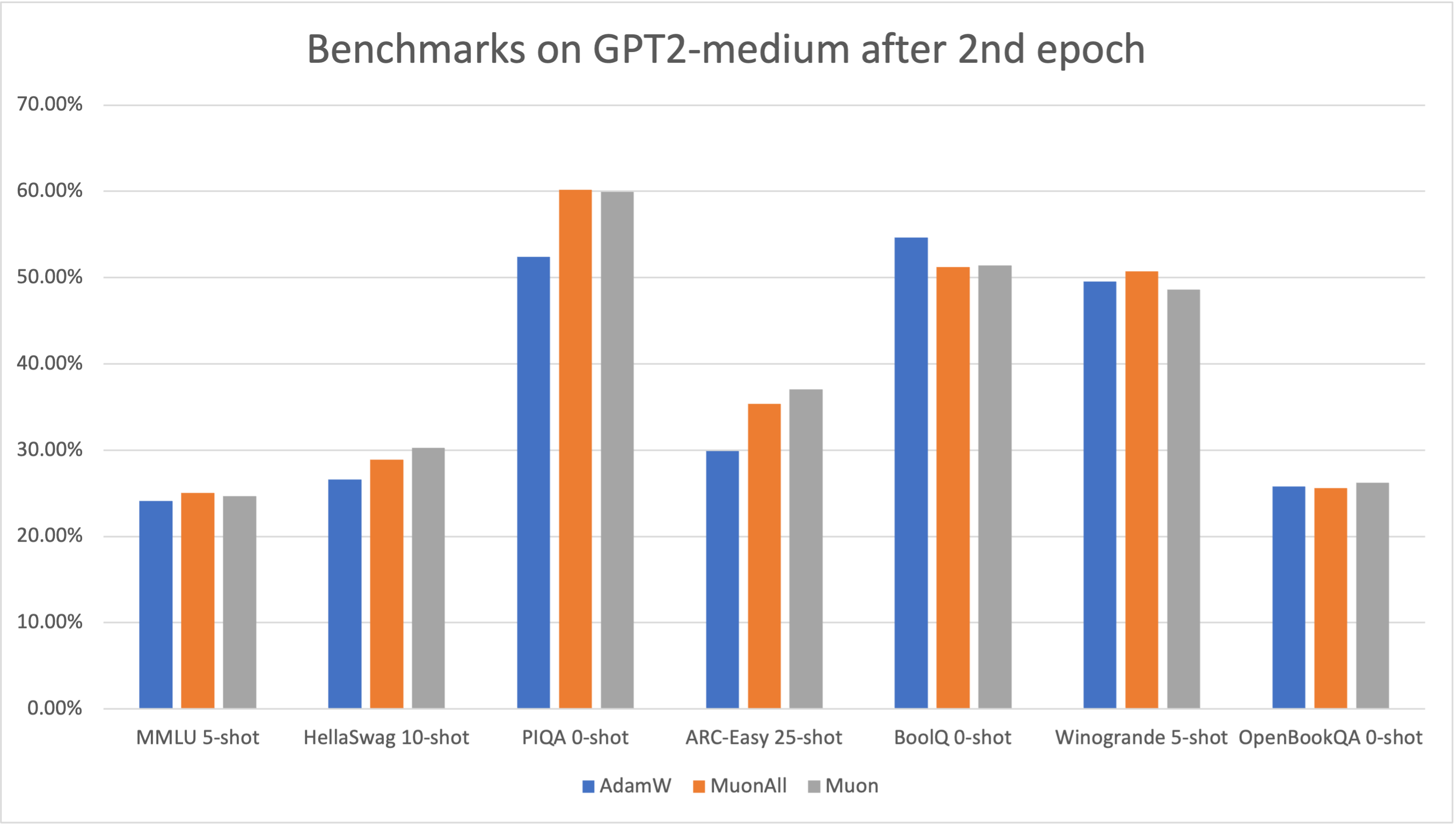} % adjust width as needed
    \caption{Benchmarks on GPT2-medium after 2nd epoch}
    \label{fig:gpt2-hist2}
\end{figure}

The performance of Muon and MuonAll is comparable with AdamW. In the benchmarks where AdamW outcompetes Muon; MuonAll outcompetes Muon as well. Muon versions perform better than AdamW in major benchmark MMLU. The figure \ref{fig:gpt2-hist2} gives a good visualization of the benchmarking scores.

\begin{table}[H]
\centering
\renewcommand{\arraystretch}{1.2}
\caption{Benchmark evaluation results of GPT2 medium}
\setlength{\tabcolsep}{3pt}
\resizebox{\linewidth}{!}{%
\begin{tabular}{l c c c c c c}
\hline
\textbf{Epoch} & \multicolumn{3}{c}{\textbf{1}} & \multicolumn{3}{c}{\textbf{2}} \\
\cmidrule(lr){2-4} \cmidrule(lr){5-7}
\textbf{Benchmark} & \textbf{Adam} & \textbf{MuonAll} & \textbf{Muon} & \textbf{Adam} & \textbf{MuonAll} & \textbf{Muon} \\
\hline
MMLU 5-shot              & 24.14\% & \textbf{25.25\%} & 24.27\% & 24.09\% & \textbf{25.05\%} & 24.65\% \\
HellaSwag 10-shot        & 26.60\% & 28.91\% & \textbf{30.22\%} & 26.62\% & 28.90\% & \textbf{30.27\%} \\
PIQA 0-shot              & 53.65\% & 60.01\% & \textbf{60.45\%} & 52.39\% & \textbf{60.17\%} & 59.90\% \\
ARC-Easy 25-shot         & 29.55\% & 35.56\% & \textbf{37.33\%} & 29.92\% & 35.35\% & \textbf{37.04\%} \\
BoolQ 0-shot             & \textbf{55.20\%} & 51.28\% & 49.30\% & \textbf{54.65\%} & 51.25\% & 51.44\% \\
Winogrande 5-shot        & 49.96\% & \textbf{50.51\%} & 48.86\% & 49.57\% & \textbf{50.75\%} & 48.62\% \\
OpenBookQA 0-shot        & \textbf{26.20\%} & 25.80\% & 25.20\% & 25.80\% & 25.60\% & \textbf{26.20\%} \\
\hline
\end{tabular}%
}
\label{tab:gpt2-bench}
\end{table}

% \begin{figure}[H]
%     \centering
%     \includegraphics[width=0.5\textwidth]{Bench_GPT_epoch1.png} % adjust width as needed
%     \caption{Benchmarks on GPT2-medium after 1st epoch}
%     \label{fig:gpt2-hist1}
% \end{figure}

% \begin{figure}[H]
%     \centering
%     \includegraphics[width=0.4\textwidth]{GPT2_EvalLoss_Latest.png} % adjust width as needed
%     \caption{Validation loss on GPT2-medium}
%     \label{fig:gpt2-val}
% \end{figure}

\begin{table}[H]
\centering
\renewcommand{\arraystretch}{1.2}
\caption{Training and validation losses}
\setlength{\tabcolsep}{8pt}
\begin{tabular}{l l c c}
\hline
\textbf{Model} & \textbf{Optimizer} & \textbf{Training loss} & \textbf{Validation loss} \\
\hline
\multirow{3}{*}{QWEN2-0.5B} 
 & AdamW     & 0.54213 & 0.52592 \\
 & MuonAll   & 0.56804 & 0.54942 \\
 & Muon      & 0.56429 & 0.55058 \\
\hline
\multirow{3}{*}{SMOLLM2-360M} 
 & AdamW     & 0.55882 & 0.54377 \\
 & MuonAll   & 0.59449 & 0.57391 \\
 & Muon      & 0.57803 & 0.56029 \\
\hline
\multirow{3}{*}{GPT2 MEDIUM} 
 & AdamW     & 0.0007  & 0.000026537 \\
 & MuonAll   & 0.6503  & 0.52772 \\
 & Muon      & 0.645   & 0.52885 \\
\hline
\end{tabular}
\label{tab:val-results}
\end{table}

% \section{Discussion}

% The preferred spelling of the word ``acknowledgment'' in America is without 
% an ``e'' after the ``g''. Avoid the stilted expression ``one of us (R. B. 
% G.) thanks $\ldots$''. Instead, try ``R. B. G. thanks$\ldots$''. Put sponsor 
% acknowledgments in the unnumbered footnote on the first page.

% Need to explore on large language models.
% - Why muon performs better in coding \& math tasks

\section{Conclusion}
This paper has addressed two possible directions that emerged after showing Muon's scalability in LLM training: (1) Can all parameters be integrated inside Muon instead of a hybrid approach using AdamW? (2) Will Muon be effective if used for SFT on already available public pretrained models? For the first question, we modify the original Muon architecture and introduce MuonAll which incorporates all the parameters in it. For the second question, we provide an extensive analysis of SFT experiments on public pretrained base models: Qwen2-0.5B, SmolLM-360M, GPT2-medium by comparing performance of AdamW, Muon and MuonAll. Muon and MuonAll finetuned models perform on par with the AdamW-finetuned model. For the benchmarks where AdamW outperforms Muon, MuonAll outperforms Muon as well. For the same number of tokens, Muon and MuonAll achieve this with a higher wall-clock time. While this work demonstrates the effectiveness of Muon for efficient and stable fine-tuning of LLMs, several aspects remain unexplored. Future research could investigate extensions of Muon beyond spectral norm constraints, exploring generalized matrix norms such as Schatten-p or trace norms to better balance convergence speed and model generalization.

\bibliographystyle{ieeetr}
\bibliography{references}

\begin{thebibliography}{10}

\bibitem{b10}
{OpenAI} {\em et~al.}, ``Gpt-4o system card,'' {\em arXiv preprint
  arXiv:2410.21276}, 2024.
\newblock [Online]. Available: https://arxiv.org/abs/2410.21276.

\bibitem{b11}
{DeepSeek-AI} {\em et~al.}, ``Deepseek-v2: A strong, economical, and efficient
  mixture-of-experts language model,'' {\em arXiv preprint arXiv:2405.04434},
  2024.
\newblock [Online]. Available: https://arxiv.org/abs/2405.04434.

\bibitem{b12}
{Gemini Team} {\em et~al.}, ``Gemini 1.5: Unlocking multimodal understanding
  across millions of tokens of context,'' {\em arXiv preprint
  arXiv:2403.05530}, 2024.
\newblock [Online]. Available: https://arxiv.org/abs/2403.05530.

\bibitem{b13}
D.~P. Kingma and J.~Ba, ``Adam: A method for stochastic optimization,'' {\em
  arXiv preprint arXiv:1412.6980}, 2017.
\newblock [Online]. Available: https://arxiv.org/abs/1412.6980.

\bibitem{b14}
I.~Loshchilov and F.~Hutter, ``Decoupled weight decay regularization,'' {\em
  arXiv preprint arXiv:1711.05101}, 2019.
\newblock [Online]. Available: https://arxiv.org/abs/1711.05101.

\bibitem{b15}
H.~Liu {\em et~al.}, ``Sophia: A scalable stochastic second-order optimizer for
  language model pre-training,'' {\em arXiv preprint arXiv:2305.14342}, 2024.
\newblock [Online]. Available: https://arxiv.org/pdf/2305.14342.pdf.

\bibitem{b1}
K.~Jordan, Y.~Jin, V.~Boza, J.~You, F.~Cesista, L.~Newhouse, and J.~Bernstein,
  ``Muon: An optimizer for hidden layers in neural networks,'' 2024.
\newblock [Online]. Available: https://kellerjordan.github.io/posts/muon/.

\bibitem{b16}
H.~Yuan {\em et~al.}, ``Mars: Unleashing the power of variance reduction for
  pretraining at scale,'' in {\em Proc. Int. Conf. Learn. Represent. (ICLR)},
  2025.
\newblock [Online]. Available: https://openreview.net/forum?id=NrcKQ3ASLZ.

\bibitem{b17}
R.~Vyas {\em et~al.}, ``Deconstructing what makes a good optimizer for
  autoregressive language modeling,'' {\em arXiv preprint arXiv:2407.07972},
  2024.
\newblock [Online]. Available: https://arxiv.org/abs/2407.07972.

\bibitem{b18}
X.-L. Li and F.~Orabona, ``On the convergence of stochastic gradient descent
  with adaptive stepsizes,'' 2018.
\newblock [Online]. Available: [URL if available].

\bibitem{b19}
X.-L. Li and F.~Orabona, ``A simple and provably convergent algorithm for
  asynchronous sgd,'' 2018.
\newblock [Online]. Available: [URL if available].

\bibitem{pooladzandi2024curvature}
O.~Pooladzandi and X.-L. Li, ``Curvature-informed sgd via general purpose
  lie-group preconditioners,'' {\em arXiv:2402.04553 [cs.LG]}, 2024.
\newblock [Online]. Available: https://arxiv.org/abs/2402.04553.

\bibitem{li2022multi}
Y.~Li, X.~Liang, J.~Liu, and H.~Zhou, ``Multi-strategy equilibrium optimizer:
  An improved meta-heuristic tested on numerical optimization and engineering
  problems,'' {\em PLoS ONE}, vol.~17, no.~10, 2022.
\newblock [Online]. Available:
  https://journals.plos.org/plosone/article?id=10.1371/journal.pone.0276210.

\bibitem{zhao2024deconstructing}
R.~Zhao, X.-L. Li, {\em et~al.}, ``Deconstructing what makes a good optimizer
  for autoregressive language modeling,'' {\em arXiv preprint
  arXiv:2407.07972}, 2024.
\newblock [Online]. Available: https://arxiv.org/abs/2407.07972.

\bibitem{pethick2025stable}
S.~Pethick {\em et~al.}, ``A stable whitening optimizer for efficient neural
  network training,'' {\em arXiv preprint arXiv:2506.07254}, 2025.
\newblock [Online]. Available: https://arxiv.org/abs/2506.07254.

\bibitem{b2}
J.~Liu {\em et~al.}, ``Muon is scalable for llm training,'' {\em arXiv preprint
  arXiv:2502.16982}, 2025.
\newblock [Online]. Available: https://arxiv.org/abs/2502.16982.

\bibitem{bernstein2024oldoptimizer}
J.~Bernstein and L.~Newhouse, ``Old optimizer, new norm: An anthology,'' {\em
  arXiv:2409.20325 [cs.LG]}, 2024.
\newblock [Online]. Available: https://arxiv.org/abs/2409.20325.

\bibitem{b6}
J.~Li and M.~Hong, ``A note on the convergence of muon,'' {\em arXiv preprint
  arXiv:2502.02900}, 2025.
\newblock [Online]. Available: https://arxiv.org/abs/2502.02900.

\bibitem{b7}
N.~Sato, H.~Naganuma, and H.~Iiduka, ``Convergence bound and critical batch
  size of muon optimizer,'' {\em arXiv preprint arXiv:2507.01598}, 2025.
\newblock [Online]. Available: https://arxiv.org/abs/2507.01598.

\bibitem{b8}
W.~Shen, R.~Huang, M.~Huang, C.~Shen, and J.~Zhang, ``On the convergence
  analysis of muon,'' {\em arXiv preprint arXiv:2505.23737}, 2025.
\newblock [Online]. Available: https://arxiv.org/abs/2505.23737.

\bibitem{b9}
L.~Chen, J.~Li, and Q.~Liu, ``Muon optimizes under spectral norm constraints,''
  {\em arXiv preprint arXiv:2506.15054}, 2025.
\newblock [Online]. Available: https://arxiv.org/abs/2506.15054.

\bibitem{b4}
C.~Si, D.~Zhang, and W.~Shen, ``Adamuon: Adaptive muon optimizer,'' {\em arXiv
  preprint arXiv:2507.11005}, 2025.
\newblock [Online]. Available: https://arxiv.org/abs/2507.11005.

\bibitem{b5}
B.~Th{\'e}rien, X.~Huang, I.~Rish, and E.~Belilovsky, ``Muloco: Muon is a
  practical inner optimizer for diloco,'' {\em arXiv preprint
  arXiv:2505.23725}, 2025.
\newblock [Online]. Available: https://arxiv.org/abs/2505.23725.

\bibitem{b3}
{Essential AI} {\em et~al.}, ``Practical efficiency of muon for pretraining,''
  {\em arXiv preprint arXiv:2505.02222}, 2025.
\newblock [Online]. Available: https://arxiv.org/abs/2505.02222.

\bibitem{qwen2}
A.~Yang, B.~Yang, B.~Hui, {\em et~al.}, ``Qwen2 technical report,'' {\em
  arXiv:2407.10671 [cs.CL]}, 2024.
\newblock [Online]. Available: https://arxiv.org/abs/2407.10671.

\bibitem{allal2025smollm2}
L.~B. Allal, A.~Lozhkov, E.~Bakouch, {\em et~al.}, ``Smollm2: When smol goes
  big—data-centric training of a small language model,'' {\em
  arXiv:2502.02737 [cs.CL]}, 2025.
\newblock [Online]. Available: https://arxiv.org/abs/2502.02737.

\bibitem{radford2019language}
A.~Radford, J.~Wu, R.~Child, D.~Luan, D.~Amodei, I.~Sutskever, {\em et~al.},
  ``Language models are unsupervised multitask learners.'' OpenAI Blog, 2019.

\bibitem{OpenOrca}
W.~Lian, B.~Goodson, E.~Pentland, A.~Cook, C.~Vong, and Teknium, ``Openorca: An
  open dataset of gpt augmented flan reasoning traces.'' HuggingFace
  repository, 2023.
\newblock [Online]. Available:
  https://huggingface.co/datasets/Open-Orca/OpenOrca.

\bibitem{hendrycks2021mmlu}
D.~Hendrycks, C.~Burns, S.~Basart, A.~Zou, M.~Mazeika, D.~Song, and
  J.~Steinhardt, ``Measuring massive multitask language understanding,'' in
  {\em Proc. Int. Conf. Learn. Represent. (ICLR)}, 2021.
\newblock [Online]. Available: https://arxiv.org/abs/2009.03300.

\bibitem{b21}
P.~Clark {\em et~al.}, ``Think you have solved question answering? try arc,''
  {\em arXiv preprint arXiv:1803.05457}, 2018.
\newblock [Online]. Available: https://arxiv.org/abs/1803.05457.

\bibitem{zellers2019hellaswag}
R.~Zellers, A.~Holtzman, Y.~Bisk, A.~Farhadi, and Y.~Choi, ``Hellaswag: Can a
  machine really finish your sentence?,'' in {\em Proc. 57th Annu. Meet. Assoc.
  Comput. Linguistics (ACL)}, 2019.

\bibitem{sakaguchi2019winogrande}
K.~Sakaguchi, R.~L. Bras, C.~Bhagavatula, and Y.~Choi, ``Winogrande: An
  adversarial winograd schema challenge at scale,'' {\em arXiv preprint
  arXiv:1907.10641}, 2019.
\newblock [Online]. Available: https://arxiv.org/abs/1907.10641.

\bibitem{b20}
Y.~Bisk, R.~Zellers, R.~L. Bras, J.~Gao, and Y.~Choi, ``Piqa: Reasoning about
  physical commonsense in natural language,'' {\em arXiv preprint
  arXiv:1911.11641}, 2019.
\newblock [Online]. Available: https://arxiv.org/abs/1911.11641.

\bibitem{b22}
C.~Clark, K.~Lee, M.-W. Chang, T.~Kwiatkowski, M.~Collins, and K.~Toutanova,
  ``Boolq: Exploring the surprising difficulty of natural yes/no questions,''
  in {\em Proc. NAACL-HLT}, (Minneapolis, MN, USA), pp.~2924--2936, 2019.
\newblock [Online]. Available: https://aclanthology.org/N19-1300.pdf.

\bibitem{cobbe2021gsm8k}
K.~Cobbe, V.~Kosaraju, M.~Bavarian, M.~Chen, H.~Jun, L.~Kaiser, M.~Plappert,
  J.~Tworek, J.~Hilton, R.~Nakano, C.~Hesse, and J.~Schulman, ``Training
  verifiers to solve math word problems,'' {\em arXiv preprint
  arXiv:2110.14168}, 2021.
\newblock [Online]. Available: https://arxiv.org/abs/2110.14168.

\bibitem{mihaylov2018openbookqa}
T.~Mihaylov, P.~Clark, T.~Khot, and A.~Sabharwal, ``Can a suit of armor conduct
  electricity? a new dataset for open book question answering,'' in {\em Proc.
  Empirical Methods Natural Lang. Process. (EMNLP)}, (Brussels, Belgium),
  pp.~2381--2391, 2018.
\newblock [Online]. Available: https://aclanthology.org/D18-1260.

\bibitem{lambadaopenai}
{OpenAI}, ``Lambada (openai variant).'' Dataset, 2024.
\newblock [Online]. Available:
  \url{https://huggingface.co/datasets/EleutherAI/lambada_openai}.

\end{thebibliography}

\end{document}